\documentclass[letter,11pt]{article}
\usepackage[utf8]{inputenc}
\usepackage{times}
\usepackage[margin=1in]{geometry}
\usepackage{natbib}
\usepackage{amsmath}
\usepackage{amssymb}
\usepackage{xcolor}
\usepackage[hidelinks]{hyperref}

\usepackage[normalem]{ulem}

\setcitestyle{authoryear,open={(},close={)}}

\title{{On Consequentialism and Fairness}}
\author{Dallas Card$^1$ \quad  Noah A. Smith$^{2,3}$ \\
\\ $^1$ Stanford University, Stanford, California, USA
\\ $^2$ University of Washington, Seattle, Washington, USA
\\ $^3$ Allen Institute for AI, Seattle, Washington, USA
}
\date{May 2020}

\begin{document}
\maketitle

\begin{abstract}
Recent work on fairness in machine learning has primarily emphasized how to define, quantify, and encourage ``fair'' outcomes. Less attention has been paid, however, to the ethical foundations which underlie such efforts. Among the ethical perspectives that should be taken into consideration is \emph{consequentialism}, the position that, roughly speaking, outcomes are all that matter. Although consequentialism is not free from difficulties, and although it does not necessarily provide a tractable way of choosing actions (because of the combined problems of uncertainty, subjectivity, and aggregation), it nevertheless provides a powerful foundation from which to critique the existing literature on machine learning fairness. Moreover, it brings to the fore some of the tradeoffs involved, including the problem of who counts, the pros and cons of using a policy, and the relative value of the distant future. In this paper we provide a consequentialist critique of common definitions of fairness within machine learning, as well as a machine learning perspective on consequentialism. We conclude with a broader discussion of the issues of learning and randomization, which have important implications for the ethics of automated decision making systems. \\
\small
%\textbf{Keywords: consequentialism, fairness, ethics, machine learning, randomization}
\end{abstract}

\maketitle

\section{Introduction}

In recent years, computer scientists have increasingly come to recognize that artificial intelligence (AI) systems have the potential to create harmful consequences. Especially within machine learning, there have been numerous efforts to formally characterize various notions of \emph{fairness} and develop algorithms to satisfy these criteria. However, most of this research has proceeded without any nuanced discussion of ethical foundations. Partly as a response, there have been several recent calls to think more broadly about the ethical implications of AI \citep{barabas.2018,hu.2018,torresen.2018,green.2019}.

Among the most prominent approaches to ethics within philosophy is a highly influential position known as \textit{consequentialism}. 
Roughly speaking, the consequentialist believes that outcomes are all that matter, and that people should therefore endeavour to \emph{act so as to produce the best consequences, based on an impartial perspective as to what is best}.

Although there are numerous difficulties with consequentialism in practice
(see \S\ref{sec:difficulties}), it nevertheless provides a clear and principled foundation from which to critique proposals which fall short of its ideals.
In this paper, we analyze the literature on fairness within machine learning, and show how it largely depends on assumptions which the consequentialist perspective reveals immediately to be problematic.
In particular, we make the following contributions:
\begin{itemize}
\item We provide an accessible overview of the main ideas of consequentialism (\S\ref{sec:definition}), as well as a discussion of its difficulties (\S\ref{sec:difficulties}), with a special emphasis on computational limitations.
\item We review the dominant ideas about fairness in the machine learning literature  (\S\ref{sec:fairness}), and provide the first critique of these ideas explicitly from the perspective of consequentialism (\S\ref{sec:perspective}). 
\item We conclude with a broader discussion of the ethical issues raised by learning and randomization, highlighting future direction for both AI and consequentialism (\S\ref{sec:algorithmic}).
\end{itemize}

\section{Motivating Examples} \label{sec:examples}

Before providing a formal description of consequentialism (\S\ref{sec:definition}), we will begin with a series of motivating examples which illustrate some of the difficulties involved.
We consider three variations on decisions about lending money, a frequently-used example in discussions about fairness, and an area in which AI could have significant real-world consequences.

First, imagine being asked by a relative for a small personal loan. 
This would seem to be a relatively low-stakes decision involving a simple tradeoff (e.g., financial burden vs. familial strife). Although this decision could in principle have massive long term consequences (perhaps the relative will start a business that will have a large impact, etc.), it is the immediate consequences which will likely dominate the decision.
On the other hand, treating this as a simple yes-or-no decision fails to recognize the full range of possibilities. 
A consequentialist might suggest that we consider all possible uses of the money, such as investing it, or lending it to someone in even greater need. 
Whereas \emph{commonsense morality} might direct us to favor our relatives over strangers, the notion of \emph{impartiality} inherent in consequentialism presents a challenge to this perspective,
thus raising the problem of \emph{demandingness} (\S\ref{sec:commonsense}). 

Second, consider a bank executive creating a policy to determine who will or will not be granted a loan.
This policy will
affect not only would-be borrowers, but also
the financial health of the bank, its employees, etc. In this case, the bank will likely be bound by various forms of regulation which will constrain the policy. Even a decision maker with an impartial perspective will be bound by these laws (the breaking of which might entail severe negative consequences). In addition, the bank might wish to create a policy that will be perceived as \emph{fair}, yet knowing the literature on machine learning fairness, they will know that no policy will simultaneously satisfy all criteria that have been proposed (\S\ref{sec:fairness}).
Moreover, there may be a tradeoff between short-term profits and long-term success (\S\ref{sec:temporal}).

Finally, consider a legislator trying to craft legislation that will govern the space of policies that banks are allowed to use in determining who will get a loan.
This is an even more high-level decision that could have even more far reaching consequences.
As a democratic society, 
we may hope that those in government will work for the benefit of all
(though this hope may often be disappointed in practice), but it is unclear how even a selfless legislator should balance all competing interests (\S\ref{sec:value}).
Moreover, even if there were consensus on the desired outcome, determining the expected consequences of any particular governing policy will be extremely difficult, 
as 
banks will react to any such legislation, trying to maximize their own interests while respecting the letter of the law,
thus raising the problem of \emph{uncertainty} (\S\ref{sec:uncertainty}).

Although these scenarios are distinct, each of the issues raised applies to some extent in each case. As we will discuss, work on fairness within machine learning has focused primarily on the intermediate, institutional case, and has largely ignored the broader context. We will begin with an in-depth overview of consequentialism that engages with these difficulties, and then show that it nevertheless provides a useful critical perspective on conventional thinking about fairness within machine learning (\S\ref{sec:perspective}).

\section{Consequentialism Defined} \label{sec:definition} 

\subsection{Overview}

The literature on consequentialism is vast, including many nuances that will not concern us here. 
The most well known expressions can be found in the writings of Jeremy \citet{bentham} and John Stuart \citet{mill}, later refined by philosophers such as Henry \citet{sidgwick}, Elizabeth \citet{anscombe.1958}, Derek \citet{parfit}, and Peter \citet{singer.1993}.
The basic idea which unifies all of this thinking is that only the \emph{outcomes} that result from our actions
(i.e., the relative value of possible worlds that might exist in the future)
have moral relevance.
%\footnote{In G. E. Moore's phrasing ``To ask what kind of actions we ought to perform, or what kind of conduct is right, is to ask what kind of effects such action and conduct will produce.'' \citep{moore}.}

Before proceeding, it is helpful to consider three lenses through which we can make sense of an ethical theory. First, we can consider a statement to be
a claim about what would be objectively best,
given some sort of full knowledge and understanding of the universe.
Second, we can think of an ethical theory as a proposed guide for how someone should choose to act in a particular situation (which may only align partially with an objective perspective, due to limited information).
Third, although less conventional, we can think of ethics as a way to interpret the actions taken by others. In the sense that ``actions speak louder than words,'' we can treat people's behavior as revealing of their view of what is morally correct \citep{greene.2002}. 

Although consequentialism is typically presented in a more abstract philosophical form (often illustrated via thought experiments), we will begin with a concise mathematical formulation of the two most common forms of consequentialism, known as \textit{act consequentialism} and \textit{rule consequentialism}.
For the moment, 
we will intentionally adopt the objective perspective, before returning to practical difficulties below. 

\subsection{Act Consequentialism}

First, consider the proposal known as \textit{act consequentialism}. This theory says, simply, that the best action to take in any situation is the one that will produce the best outcomes \citep{smart.1973,railton}.
To be precise, let us define the set of possible actions, $\mathcal{A}$, and an evaluation function $v(\cdot)$. 
According to act consequentialism, the best action to take is the one that will lead to the consequences with the greatest value,
i.e., 
\begin{equation} \label{eq:nonrandom_consequantialism}
    a^{*} = \underset{a \in \mathcal{A}}{\arg \max}~v(c_a),
\end{equation}
where $v(c_a)$ computes the value of consequences, $c_a$, which follow from taking action $a$. Importantly, note that $c_a$ here represents not just the local or immediate consequences of $a$, but \textit{all} consequences \citep{kagan.1998}. In other words, we can think of the decision as a branching point in the universe, and want to evaluate how it will unfold based on the action that is taken at a particular moment in time \citep{portmore.2011}.

While Eq.~\ref{eq:nonrandom_consequantialism} might seem tautological, it is by no means a universally agreed upon definition of what is best. For example, many \emph{deontological} theories posit that certain actions should never be permitted (or that some might always be required), no matter what the consequences. In addition, there are some obvious difficulties with Eq.~\ref{eq:nonrandom_consequantialism}, especially the question of how to define the evaluation function $v(\cdot)$. We will return to this and other difficulties below (\S\ref{sec:difficulties}), but for the moment we will put them aside. 

One might object that perhaps there is inherent randomness in the universe, leading to uncertainty about $c_a$. In that case, we can sensibly define the optimal action in terms of the expected value of all future consequences, i.e., 
\begin{equation} \label{eq:act_consequentialism}
    a^{*} = \underset{a \in \mathcal{A}}{\arg \max}~ \mathbb{E}_{p(c \mid a)} [v(c)],
\end{equation}
where $p(c \mid a)$ represents the true probability (according to the universe) that consequences $c$ will follow from action $a$. That is, for each possible action, we would consider all possible outcomes which might result from that action, and sum their values, weighted by the respective probabilities that they will occur, recommending the action with the highest expected value.

To make the dependence on future consequences more explicit, it can be helpful to factor the expected value into a summation over time, optionally with some sort of discounting. Although consequentialism does not require that we factorize the value of the future in this way, it will prove convenient in further elaboration of these ideas.
%\footnote{In particular, temporal aggregation \emph{within} a human life presents unique challenges (see \citep{parfit}), but a simple summation with discounting is arguably a reasonable formulation for longer time scales.}
For the sake of simplicity, we will assume that time can be discretized into finite steps. A statement of act consequentialism using a simple geometric discounting factor would then be:
\begin{equation} \label{eq:act_consequentialism_time}
    a^{*} = \underset{a \in \mathcal{A}}{\arg \max}~ \sum_{t=0}^\infty \gamma^{t} \cdot \mathbb{E}_{p(s_{t+1} \mid a)}[v(s_{t+1})],
\end{equation}
where $p(s_{t+1} \mid a)$ represents the probability that the universe will be in state $s$ at time $t+1$ if we take action $a$ at time $t=0$, and $0 \leq \gamma \leq 1$ represents the discount factor.
A discount factor of $0$ means that only the \textit{immediate} consequences of an action are relevant, whereas a discount factor of $1$ means that all times in the future are valued equally.\footnote{One could similarly augment Eq. \ref{eq:act_consequentialism_time} to make any epistemic uncertainty about the evaluation function or discount factor explicit.} 

\subsection{Rule Consequentialism}

The main alternative to act consequentialism is a variant known as \emph{rule consequentialism} \citep{harsanyi.1977,hooker.2002}.
As the name suggests, rule consequentialism is similar to act consequentialism, except that rather than focusing on the best action in each unique situation, it suggests that we should act according to a set of \emph{rules} governing all situations, and 
adopt the set of rules which will lead to the best overall outcomes.\footnote{In some cases, rule consequentialism is formulated as the problem of choosing the set of rules which, if internalized by the vast majority of the community, would lead to the best consequences \citep{hooker.2002}.}
%\footnote{Note that formulations of rule consequentialism typically include the possibility of requiring different behaviour from different people based on their circumstances, such as the idea that wealthier people should give a larger proportion of their income to charity, which can alternatively be viewed as different rules for different people \citep{kagan.1998}.}

Here, we will refer to a set of rules as a \textit{policy}, and allow for the policy to be stochastic. In other words, a policy, $\pi$, is a probability distribution over possible actions conditional on the present state $s$, i.e., $\pi(s) \triangleq p(a \mid s)$. To make a decision, an action is sampled randomly from this distribution.\footnote{Most treatments of consequentialism assume that the rules determine a single correct action for each situation. However, the formulation presented here is strictly more general;  deterministic policies are those that assign all probability mass to a single action for each state.}
Using the same temporal factorization as above, we can formalize rule consequentialism as
\begin{equation} \label{eq:rule_consequentialism}
    \pi^{*} = \underset{\pi \in~\Pi}{\arg \max}~  \mathbb{E}_{p(s_{t'+1} \mid a_{t'}, s_{t'}) \pi(a_{t'} \mid s_{t'})} \left[ \sum_{t=0}^\infty \gamma^t \cdot v(s_{t+1}) \right], 
\end{equation}
where $\Pi$ represents the space of possible policies, and the expectation is now taken with respect to the governing dynamics, in which actions are selected based on the state of the world, i.e. $a_t \sim \pi(a_t \mid s_t)$, and the next state depends on the current state of the world and the action taken, i.e., $s_{t+1} \sim p(s_{t+1} \mid s_t, a_t)$. 

While some have suggested that rule consequentialism is strictly inferior to act consequentialism, in that it fails to treat each situation as unique \citep{railton}, others have argued for it, citing the inability of individuals to accurately determine the best action in each unique situation \citep{hooker.2002}, as well as benefits from coordination and incentives \citep{harsanyi.1977}.
As noted by various papers (e.g., \citealp{abel.2016}), Eq.~\ref{eq:rule_consequentialism} bears a striking resemblance to the problem of \emph{reinforcement learning}.\footnote{Eq.~\ref{eq:rule_consequentialism} is equivalent to the standard formulation of a Markov decision process if we restrict ourselves to a finite set of states $s \in \mathcal{S}$, actions $a \in \mathcal{A}$, transition probabilities $p(s_{t+1} \mid s_t, a_t)$, and discount factor $\gamma$.}
While this similarity is provocative, we will defer discussion of it (and the more general question of \emph{learning}) until section \S\ref{sec:algorithmic}.

It is important to emphasize that the above formulation is a highly stylized discussion of morality, largely divorced from reality, which tries to encapsulate a large body of philosophical writing put forward under the name  ``consequentialism''. Thinking about what this formulation has to tell us about how individuals make (or should make) choices requires further elaboration, which we revisit below (\S\ref{sec:difficulties}).

\subsection{Competing Ethical Frameworks}

The primary contrasting proposals to consequentialism are a) \emph{deontology}; and b) theories in the \emph{social contract} tradition.  As mentioned above, deontological theories posit that there are certain restrictions or requirements on action, \textit{a priori}, which cannot be violated. For example, various religious traditions place restrictions on lending money, or require a certain level of charitable giving.
Using the framework established above, we can describe deontological theories as \emph{constraints} on the action space, $\mathcal{A}$, or policy space, $\Pi$ \citep{kagan.1998}.
While they may accord more with our commonsense notions of morality (see \S\ref{sec:commonsense}), deontological theories are open to challenge because of their inability to justify the particular constraints they specify, as well as the implication that they would fail to produce the best outcomes in certain scenarios \citep{scheffler,smart.1973}.

By contrast, social contract theories are more concerned with determining the rules, or ways of organizing society, that a group of free and reasonable people would agree to in an idealized deliberative scenario.\footnote{E.g., ``An act is wrong if its performance under the circumstances would be disallowed by any set of principles for the general regulation of behaviour that no one could reasonably reject as a basis for informed, unforced, general agreement'' \citep{scanlon.1998}.} Most famously in this tradition, John Rawls suggested
that we should imagine people designing society behind a ``veil of ignorance'', not knowing what position they will hold in that society \citep{rawls.1971}. We cannot possibly do justice to these other schools of thought in the space available, but we note that there is value in thinking about sociotechnical systems from multiple ethical perspectives, and encourage others to elaborate on these points.\footnote{For a review of how Rawls has been applied within information sciences, see \citet{hoffmann.2017}.}

In this paper, we focus on consequentialism not because it is necessarily superior to the alternatives, but because it is influential, and because it might seem, at first glance, to have a natural affinity with machine learning and optimization.
%\footnote{In a recent survey of professional philosophers, 24\% responded that they ``accept or lean toward consequentialism'' \citep{bourget.2014}.}
While there have been many papers providing brief summaries of various ethical theories and their relevance to AI, we believe that a more in-depth treatment is required to fully unpack the implications of each, and would encourage similar consideration of the above traditions, as well as virtue ethics, feminist ethics, etc.

Before discussing the problems with consequentialism, it is useful to note that the formulation given in Eq. \ref{eq:rule_consequentialism} highlights three important matters about which reasonable people might disagree, with respect to how we should act (alluded to in \S\ref{sec:examples}): we might disagree about the relative value of different outcomes (the evaluation function, $v(\cdot)$); we might disagree about the likely effects of different actions (the probability of outcomes, $p(s_{t+1} \mid s_t, a_t)$); and we might disagree about how much weight to place on the  distant future (the discount factor, $\gamma$).

\section{Difficulties of Consequentialism} \label{sec:difficulties}

Even if one accepts the idea in Eq. \ref{eq:act_consequentialism}---that the best action is the one that will produce the best outcome in expectation, with no \textit{a priori} restrictions on the action space, there are still numerous difficulties with consequentialism, both theoretically and in practice. 

\subsection{Value} \label{sec:value}

Perhaps the most vexing part of consequentialism is the evaluation function, $v(\cdot)$. Even if one had perfect knowledge of how the universe would unfold conditional on each possible action, choosing the \emph{best} action would still require some sort of objective way of characterizing the relative value of each possible outcome. Most writers on consequentialism agree that the specification of value should be \emph{impartial}, in that it should not give arbitrary priority to particular individuals \citep{singer.1993,kagan.1998},
but this is far from sufficient for resolving this difficulty.\footnote{\citet{sidgwick} writes, ``I obtain the self-evident principle that the good of any one individual is of no more importance, from the point of view (if I may say so) of the Universe, than the good of any other; unless, that is, there are special grounds for believing that more good is likely to be realized in the one case than in the other.''} 
%\footnote{Rawls provides the following description of the impartial spectator: ``Endowed with ideal powers of sympathy and imagination, the impartial spectator is the perfectly rational individual who identifies with and experiences the desires of others as if these desires were his [sic] own.'' \citep{rawls.1971}.}

By far the most common way of simplifying the evaluation of outcomes, both within writings on consequentialism and in decision theory, is to adopt the classic \textit{utilitarian} perspective \citep{mill,smart.1973}. Although there are many variations, the most common statement of utilitarianism is that the value of a state is equal to the sum of the well-being experienced by all individual entities.\footnote{The philosophical literature in some cases uses happiness or the satisfaction of preferences, rather than well-being, but this distinction is not essential for our purposes.}
%\footnote{Philosophers debate how to characterize value (happiness, suffering, fulfillment of preferences, etc.) as well as which entities (e.g. humans, animals, or all sentient beings) should count in this calculation, but these distinctions are not essential for our discussion of this issue.}
The most common social welfare function is thus
\begin{equation} \label{eq:utilitarianism}
v(s) = \sum_{e \in \mathcal{E}} w_e(s),
\end{equation}
where $\mathcal{E}$ represents the set of entities under consideration, and $w_e(s)$ measures the absolute well-being of entity $e$ in state $s$.\footnote{Note that using a separate value function for each entity accounts for variation in preferences, and allows for some entities to ``count'' for more than others, as when the set of relevant entities includes animals, or all sentient beings \citep{kagan.1998}.}
%\footnote{We might hope that there could be a single value function $w_*(s)$ applicable to all entities, but it seems impossible to avoid having multiple well-being functions, $w_e(s)$, both because different individuals may have different preferences, and because different types of entities might ``count'' for more than others. For example, while one particular formulation of value might consider only the well-being of humans, others have argued for including a larger set of entities, such as animals, or all sentient beings \citep{kagan.1998,singer.1993}.}

Although utilitarianism is highly influential, there are fundamental difficulties with it. First, aggregating well-being requires \emph{measuring} individual welfare, but it is unclear that it can be measured in a way that allows for fair comparisons, at least given current technology. Even if we restrict the set of morally relevant entities to humans, issues of subjectivity, disposition, and self-reporting make it difficult if not impossible to meaningfully compare across individuals \citep{binmore.2009}.

Second, even if there were a satisfactory way of measuring individual well-being, there are computational difficulties involved in \emph{estimating} these values for hypothetical worlds. Given that well-being could depend on fine-grained details of the state of the world, it is unclear what level of precision would be required of a model in order to evaluate well-being for each entity. Thus, even estimating the overall value of a single state of the world might be infeasible, let alone a progression of them over time.

Third, any function which maps from the welfare of multiple entities to a single scalar will fail to distinguish between dramatically different distributions. Using the sum, for example, will treat as equivalent two states with the same total value, but with different levels of inequality \citep{parfit}. While this failing is not necessarily insurmountable, most solutions seem to undermine the inherent simplicity of the utilitarian ideal.\footnote{For example, one could model well-being as a non-linear, increasing, concave (e.g. logarithmic) function of other attributes such as wealth (i.e., diminishing marginal utility), which would encourage a more equal distribution of  resources. Alternatively, one could try to incorporate people's suffering due to inequality into their value functions \citep{singer.2017}.
}

Fourth, others have challenged the ideal of impartiality on the grounds that it is subtly paternalist, emphasizes individual autonomy over relationships and care, and ignores existing relations of power \citep{friedman.1991,smart.1973,driver.2005,kittay.2009}.
Undoubtedly, there is a long and troubling history of otherwise enlightened philosophers presuming to know what is best for others, and being blind to the harms of institutions such as colonialism, while believing that certain classes of people either don't count or are incapable of full rationality \citep{mills.1997,schultz.2005}. 

Ultimately, it seems inescapable to conclude that there is no universally acceptable evaluation function for consequentialism. Rather, we must acknowledge that every action will entail an uneven distribution of costs and benefits. Even in the case where an action literally makes everyone better off, it will almost certainly benefit some more than others. As such, the most credible position is to view the idea of valuation (utilitarian or otherwise) as inherently contested and political. While we might insist that an admissible evaluation function conform to certain criteria, such as disinterestedness, or not being self-defeating \citep{parfit}, we must also acknowledge that advocating for a particular notion of value as correct is fundamentally a political act.

\subsection{Temporal Discounting} \label{sec:temporal}

Even if there were an unproblematic way of assessing the relative value of a state of the world, the extent to which we should value the distant future is yet another point of potential disagreement. It is common (for somewhat orthogonal reasons) to apply temporal discounting in economics, but it is not obvious that there is any good reason to do so when it comes to moral value \citep{cowen.1992,cowen.2006}. Just as philosophers such as Peter Singer have argued that we should not discount the value of a human life simply because a person happens to live far away \citep{singer.1972}, one could argue that the lives of those who will live in the future should count for as much as the lives of people who are alive today.  

Unfortunately, it is difficult to avoid discounting in practice, 
as it becomes increasingly difficult to predict the consequences of our actions farther into the future. Even if we assume a finite action space, the number of possible worlds to consider will grow exponentially over time. Moreover, because of the chaotic nature of complex systems, even if we had complete knowledge of the causal structure of the universe, we would be limited in our ability to predict the future by lack of precision in our knowledge about the present. 

Despite these difficulties, consequentialism would suggest that we should, to the extent that we are able, think not only about the immediate consequences of our actions, but about the longer-term consequences as well \citep{cowen.2006}. Indeed, considering the political nature of valuation, we arguably bear even greater responsibility for thinking about future generations than the present, given that those who have not yet been born are unable to directly advocate for their interests. 

\subsection{Uncertainty} \label{sec:uncertainty}

In practice, of course, we do not know with any certainty what the consequences of our actions will be, especially over the long term. 
Again, from the perspective of determining the objectively morally correct action, one might argue that all that matters is the (unknown) probability according to the universe. For individual decision makers, however, any person's ability to predict the future will be limited, and, indeed, will likely vary across individuals. In other words, it is not just our uncertainty about consequences that is a problem, but our uncertainty about our uncertainty: we don't know how well or poorly our own model of the universe matches the true likelihood of what will happen \citep{kagan.1998,cowen.2006}.
%\footnote{Even Bentham raised this issue in informal terms: ``It is not to be expected that this process should be strictly pursued previously to every moral judgement, or to every legislative or judicial operation. It may, however, be always kept in view: and as near as the process actually pursued on these occasions approaches to it, so near will such process approach to the character of an exact one'' \citep{bentham}.}

The \emph{subjective} interpretation of consequentialism suggests that, regardless of what the actual consequences may be, the morally correct thing for an individual to do is whatever they have reason to believe will produce the best consequences \citep{kagan.1998}. 
This, however, is problematic for two reasons: first, it ignores the computational effort involved in trying to determine which action would be best (which is itself a kind of action); and second, it seemingly absolves people from wrong-doing who happen to have a poor model of the world.

Rule consequentialism arguably provides a (philosophical) solution for these problems, in that it involves a direct mapping from states to actions, without requiring that each decision maker independently determine the expected value of each possible action \citep{kagan.1998,hooker.2002}.\footnote{To use a somewhat farcical example, we could imagine using a neural network to map from states to actions; the time to compute what action to take would therefore be constant for any scenario.} It still has the problem, however, of determining what policy is optimal, given our uncertainty about the world. Nevertheless, we should not overstate the problem of uncertainty; we are not in a state of total ignorance, and in general, trying to help people is likely to do more good than trying to harm them \citep{singer.2017}.

\subsection{Conflicts with Commonsense Morality} \label{sec:commonsense}

A final set of arguments against consequentialism take the form of thought experiments in which consequentialism (and utilitarianism in particular) would seemingly require us to take actions that violate our own notions of commonsense morality. A particularly common example is the ``trolley problem'' and its variants, in which it is asked whether or not it is correct to cause one person to die in order to save multiple others \citep{foot.1967,greene.2013}. 

We will not dwell on these thought experiments, except to note that many of the seeming conflicts from this type of scenario vanish once we take a longer term view, or adopt a broader notion of value than a simple sum over individuals.
Killing one patient to save five might create greater aggregate well-being if we only consider the \emph{immediate} consequences. If we consider \emph{all} consequences of such an action, however, it should be obvious why we would not wish to adopt such a policy \citep{kagan.1991}.

It is worth commenting, however, on one particular conflict with commonsense morality, namely the claim that consequentialism is, in some circumstances, excessively \emph{demanding}. Given the present amount of suffering in the world, and the diminishing marginal utility of wealth, taking consequentialism seriously would seem to require that we sacrifice nearly all of our resources in an effort to improve the well-being of the worst off \citep{smart.1973,driver.2012}. While to some extent this concern is mitigated by the same logic as above (reducing ourselves to ruin would be less valuable over the long term than sacrificing a smaller but sustainable amount), we should take seriously the possibility that the best action might not agree with our moral intuitions.

\section{Fairness in Machine Learning} \label{sec:fairness}

With the necessary background on consequentialism in place, we now review and summarize ideas about fairness in machine learning.
Note that ``fairness'' is arguably an ambiguous and overloaded term in general usage; 
our focus here is on how it has been conceptualized and formalized within the machine learning literature.\footnote{Extensive discussion of the idea of fairness can be found in much of the philosophical and technical literature cited throughout. In particular, we refer to the reader to \citet{rawls.1958}, \citet{kagan.1998}, and \citet{binnis.2018}.} In order to lay the foundation for a critical perspective on this literature, we first summarize the general framework that is commonly used for discussing fairness, and then summarize the most prominent ways in which it has been defined.\footnote{While there is also some work on fairness in the unsupervised setting (e.g. \citealp{kleindessner.2019}; \citealp{benthall.2019}), in this paper we focus on the supervised case.}

%While some might equate fairness with strict equality, others relate it to the legitimate outcome of a just process. Rawls, for example, is clear to admit the possibility of a certain amount of productive inequality (as might arise through competition, which may itself have larger benefits), suggesting that the usual sense of justice is ``the elimination of arbitrary distinctions and the establishment, within the structure of a practice, of a proper balance between competing claims'' \citep{rawls.1958}. 

The typical setup is to assume that there are two or more groups of individuals which are distinguished by some ``protected attribute'', $A$, such as race or gender. All other information about each individual is  represented by a feature vector, $X$.
The purpose of the system is to make a prediction about each individual, $\hat Y$, which we will assume to be binary, for the sake of simplicity. Moreover, we will assume that the two possible predictions (1 or 0) are asymmetric, such that one is in some sense preferable. Finally, we assume that, for some individuals, we can observe the true outcome, $Y$. We will use $\mathcal{X}$ to refer to a set of individuals.

To make this more concrete, consider the case of deciding whether or not to approve a loan. 
An algorithmic decision making system would take the applicant's information ($X$ and possibly $A$), and return a prediction about whether or not the applicant will repay the loan, $\hat Y$.
For those applicants who are approved, we can then check to see who actually pays it back on time ($Y=1$) and who does not ($Y=0$). Note, however, that in this setup, we are unable to observe the outcome for those applicants who are denied a loan, and thus cannot know what their outcome would have been in the counterfactual scenario. 

The overriding concern in this literature is to make predictions that are highly accurate while respecting some notion of fairness. 
Because reducing complex social constructs such as race and gender to simplistic categories is inherently problematic, as a running example we will instead use \textit{biological age} as a hypothetical protected attribute.\footnote{Age is a particularly interesting example of a protected attribute, as it is explicitly used to discriminate in some domains (as in restricting the right to vote), but afforded some protections in others (such as the U.S.  Age Discrimination in Employment Act).} Using the same notation as above, we would say that an automated system instantiates a policy, $\pi$, in making a prediction for each applicant. Thus, for instance $i$, a threshold classifier would predict
\begin{equation}
\hat y_i = \underset{y \in \{0, 1\}}{\arg \max}~ \pi(Y=y \mid X=x_i, A=a_i),
\end{equation}
though we might equally consider a randomized predictor.

Much of the work in fairness has drawn inspiration from two legal doctrines: \textit{disparate treatment} and \textit{disparate impact} \citep{ruggieri.2010,barocas.2016}. Disparate treatment, roughly speaking, says that two people should not be treated differently if they differ only in terms of a protected attribute. For our running example, this would be equivalent to saying that one cannot deny someone a loan simply because of their age.

Disparate impact, on the other hand, prohibits the adoption of policies that would have consequences that are unevenly distributed according to the protected attribute, even if they are neutral on their face. Thus a policy which denies loans to people with no credit history might have a disparate impact on younger borrowers, and could therefore (hypothetically) be considered discriminatory.

While research in machine learning fairness is ongoing, most proposals can be classified into two types, which to some extent map onto the two legal doctrines mentioned above. Some definitions are specified without reference to outcomes (\S\ref{sec:fairness-nonoutcomes}). Others are specified exclusively with regard to a particular set of outcomes (which must be evaluated using real data; \S\ref{sec:fairness-outcomes}). We summarize the dominant proposals of each type below.
%\footnote{Note that these categories are generally aligned with, but distinct from the more common categories of individual and group-based fairness metrics \citep{dwork.2012}.}

\subsection{Fairness Constraints Specified without Regard to Outcomes} \label{sec:fairness-nonoutcomes}

The first type of approach to fairness advocates constraints that are specified without reference to actual effects. In a formal sense, we can think of these as placing restrictions, \textit{a priori}, on the space of policies which will be considered morally acceptable. We provide three examples of this type of approach below.

\textbf{Fairness through unawareness}: A commonsense but naive notion is to disallow policies which use the protected attribute in making a prediction. Equivalently, this requires that for any $x$,
\begin{equation} \label{eq:unawareness}
\pi(y \mid x, A=0) = \pi(y \mid x, A=1)
\end{equation}
Although this seems like a strict translation of the prohibition against disparate treatment, it is generally considered to be unhelpful \citep{hardt.2016,kleinberg.2018}. Due to correlations, it may be possible to infer the protected attribute from other features, hence prohibiting a single piece of information may have no effect in practice. 

\textbf{Individual fairness}: A more general application of the same idea argues that models must make similar predictions for similar individuals  (in terms of their representations, $X$) \citep{dwork.2012}. This proposal was originally framed as being in the Rawlsian tradition, suggesting it should be a matter of public deliberation to determine who counts as similar. However, as has been noted, the effects of this framework are highly dependent on the particular notion of similarity that is chosen \citep{green.2018}. %\footnote{As Green and Hu  remark, ``The word `similar' is capricious enough to account for almost any disagreement people may have about the substantive demands of fairness'' \citep{green.2018}.}

\textbf{Randomization}: A further way of avoiding disparate treatment is through randomization \citep{kroll.2017}. The basic idea is that a policy should not look at the protected attribute \emph{or any other attribute} when making a decision, except perhaps to verify that some minimal criteria are met. For example, a policy might assign 0 probability to instances that do not meet the criteria, and  an equal probability to all others.
%, i.e.
%\begin{align}
%p(Y=1 \mid x_i \in \mathcal{X'}) &= 1/|\mathcal{X'}|\\ %\frac{1}{|\mathcal{X}|} \\
%p(Y=1 \mid x_i \notin \mathcal{X'}) &= 0,
%\end{align}
%where $\mathcal{X'}$ represents the set of individual which meet the minimal criteria (e.g., those of voting age or older). %, and $c \geq 1$.
%In expectation, this policy would approve $\frac{|\mathcal{X}|}{c}$ applicants.
Although this is a severe limitation on the space of policies, we do see instances of it being used in practice, such as in the U.S. Diversity Visa Lottery \citep{perry.2015,kroll.2017}.\footnote{Additional examples of randomization include jury selection, military service, sortition in ancient Athenian government, and which members of a firing squad have guns with real bullets. Of course, as \citet{kroll.2017} point out, randomization is only fair if the system cannot be manipulated by either applicants or decision makers.}

\subsection{Fairness Constraints Specified in Terms of Outcomes} \label{sec:fairness-outcomes}

The other major approach to fairness in machine learning is to specify requirements on the actual outcomes of a policy. In other words, while the above fairness criteria can be evaluated without data, the following criteria can only be checked using an actual dataset.
These notions of fairness are often justified in terms of the doctrine of disparate impact -- that is, policies should not be adopted which have adverse outcomes for protected groups.
Three examples are presented below:

\textbf{Demographic/statistical parity}: The notion of parity implies that the proportion of predicted labels should be the same, or approximately the same for each group. For example, this might require that an equal proportion of older and younger applicants would receive a loan. Formally, this requirement says that in order to be acceptable, a policy must satisfy
\begin{equation} \label{eq:parity}
\frac{ \sum_{i \in \mathcal{X}} \mathbb{I}[a_i = 0] \cdot \hat y_i}{ \sum_{i \in \mathcal{X}} \mathbb{I}[a_i = 0]} = \frac{ \sum_{j \in \mathcal{X}} \mathbb{I}[a_j = 1] \cdot \hat y_j}{ \sum_{j \in \mathcal{X}} \mathbb{I}[a_j = 1]},
\end{equation} 
where $\mathbb{I}[\cdot]$ equals 1 if the condition holds (otherwise 0). 
Demographic parity is a strong statement about what the consequences of a policy must be (in terms of a very focused set of short-term consequences). Note, however, that enforcing this constraint may result in suboptimal outcomes from the perspective of other criteria \citep{corbett.2017}.

\textbf{Equality of odds/opportunity}: Another outcome-based fairness criteria looks at the outcomes that result from the policy, and compares the rates of true positives and/or false positives among a held-out dataset \citep{hardt.2016}. Equal opportunity would require that, for example, an equal proportion of applicants from each group \emph{who will pay back a loan} are in fact approved. Formally, 
\begin{equation}
\frac{ \sum_{i \in \mathcal{X}} \mathbb{I}[a_i = 0, y_i=1] \cdot \hat y_i}{ \sum_{i \in \mathcal{X}} \mathbb{I}[a_i = 0, y_i=1]} = \frac{ \sum_{j \in \mathcal{X}} \mathbb{I}[a_j = 1, y_j=1] \cdot \hat y_j}{ \sum_{j \in \mathcal{X}} \mathbb{I}[a_j = 1, y_j=1]},
\end{equation} 
Equality of odds is similar, except that is requires that rates of both true positives and false positives be the same across groups.

\textbf{Equal calibration}: An alternative to equality of odds is to ask that the predictions be equally well calibrated across groups. %\footnote{Equal calibration has also been called ``test fairness'' \citep{chouldechova.2016}.}
That is, if we bin the predicted probabilities into a set of bins, a well-calibrated predictor should predict probabilities such that the proportion of instances that are correctly classified within each bin is the same for all groups. In other words, equal calibration tries to ensure that
\begin{equation}
\frac{\sum_{i \in \mathcal{X}} \mathbb{I}[a_i = 0, \hat p_i \in [b, c)] \cdot y_i}{\sum_{i \in \mathcal{X}} \mathbb{I}[a_i = 0, \hat p_i \in [b, c)]} = \frac{\sum_{j \in \mathcal{X}}  \mathbb{I}[a_j = 1, \hat p_j \in [b, c)] \cdot y_j}{\sum_{j \in \mathcal{X}} \mathbb{I}[a_j = 1, \hat p_j \in [b, c)]} 
\end{equation} 
for each interval $[b, c)$, where $\hat p_i = \pi(Y=1 \mid x_i, a_i)$ according to the policy.

Note that whereas demographic parity only requires the set of predictions ($\hat Y$) made for all individuals in a dataset, equal opportunity and equal calibration also require that we know the true outcome ($Y$) for all such individuals, even those who are given a negative prediction. As a result, the latter two requirements can only be properly verified on a dataset for which we can independently observe the true outcome (e.g., based on  assigning treatment randomly).

As has been shown by multiple authors, certain fairness criteria will necessarily be in conflict with others, under mild conditions, indicating that we will be unable to satisfy all simultaneously \citep{chouldechova.2017,kleinberg.2017}.

\section{A Consequentialist Perspective on Machine Learning Fairness} \label{sec:perspective}

As previously mentioned, most fairness metrics have been proposed with only limited discussion of ethical foundations. In this section, we provide commentary on the criteria described above from the perspective of consequentialism. As a reminder, we are not suggesting that consequentialism provides the last word on what is morally correct. Rather, we can think of consequentialism as providing one of several possible ethical perspectives which should be considered.

First, consider the fairness proposals that are specified without regard to outcomes (\S\ref{sec:fairness-nonoutcomes}). As mentioned above, these can be seen as restrictions on the set of policies that are acceptable. By definition, these constraints are not determined by the actual consequences of adopting them, nor do they possess an in-built verification mechanism to assess the nature of the consequences being produced. As such, these have more of a deontological flavor, reflecting a prior stipulation that similar people should be treated similarly, or that everyone deserves an equal chance. For example, Eq. \ref{eq:unawareness} specifies precisely the constraint on the policy space required by fairness through unawareness, and similarly for the other proposals.
In principle, of course, these criteria could have been developed with the expectation that using them would produce the best outcomes, but it is far from obvious that this is the case.

By contrast, the fairness criteria specified explicitly in terms of outcomes (\S\ref{sec:fairness-outcomes}) might seem to be closer to a form of consequentialism, given that they are evaluated by looking at actual impacts.
However, upon closer inspection we see that they imply a severely restricted form of consequentialism in terms of how they think about value, time horizon, and who counts.
In particular, while the proposals differ in terms of the precise values that are being emphasized, all of these proposals have some features in common:
\begin{itemize}
\item they only evaluate outcomes in terms of the people who are the direct object of the decision being made, not others who may be affected by these decisions;
\item they only explicitly consider the immediate consequences of each decision, equivalent to using a discount factor of 0;
\item they presuppose that a particular function of the distribution of predictions and outcomes (e.g., calibration) is the only value that is morally relevant.
\end{itemize}

Again, it is entirely possible that these constraints were developed with the \emph{intention} of producing more broadly beneficial consequences over the long term. The point is that there is nothing in the constraints themselves that points to or tries to verify this broader impact, despite the fact that they are evaluated in terms of (a narrow set of) outcomes.

To make this concrete, consider again the case of trying to regulate algorithms which will be used by banks in making loans.  Requiring satisfaction of any of the above fairness constraints will alter the set of loan applicants who are approved (and denied). While it is possible that some of these criteria might lead to broadly beneficial changes (e.g., demographic parity might enhance access to credit among those who have been historically marginalized), from the perspective of consequentialism it insufficient to evaluate the outcome only in terms of the probabilities or labels assigned to each group. Rather, it is necessary to consider the full range of consequences to individuals and society. In some cases, a loan might positively transform a person's life, or the life of their community, via mechanisms such as education and entrepreneurship. In other cases, easier access to credit could lead to speculative borrowing and financial ruin. For example, while not directly related to concerns about fairness, the potentially devastating effects of lending policies which ignore long-term and systemic effects can easily be seen in the aftermath of the subprime mortgage crisis, which derived, in part, by perverse incentives and risky lending \citep{bianco.2008}. 

Crafting effective financial regulation is obviously extremely difficult, and this is not meant to suggest that any particular fairness constraint is likely to lead to disaster. Nevertheless, it is important to remember that fairness criteria which are specified only in terms of a narrow set of short term metrics do not guarantee positive outcomes beyond what they measure, and may in some cases lead to overall greater harm.

In sum, adopting a consequentialist perspective reveals numerous ways in which the existing proposals for thinking about fairness in machine learning are fatally flawed. While all have their merits, none have been adequately justified in terms of their likely consequences, broadly considered.  Moreover, most are highly restricted in terms of the types of outcomes they take into consideration, and largely ignore broader systemic effects of adopting a single policy. 

It is, of course, understandable that most approaches to machine learning fairness have focused on \emph{a priori} constraints and tractable short term consequences. Avoiding negative consequences from new technologies is challenging in general, and many of the difficulties of consequentialism also apply directly to machine learning, especially in social contexts (uncertainty about the future, lack of agreement about value, etc.). 
Even in relatively controlled environments, it is easy to find examples of undesirable outcomes resulting from ill-specified value functions, improper time horizons, and the kinds of computational difficulties described in section \S\ref{sec:difficulties} \citep{amodei.2016}. 

Although consequentialism does not provide any easy answers about how to make AI systems more fair or just, several important considerations follow from its tenets. First, consequentialism reminds us of the need to consider outcomes broadly; technical systems are embedded in social contexts, and policies can have widespread effects on communities, not merely those who are subject to classification. Second, the political nature of valuation means that a broad range of perspectives on what is desirable should be sought out and considered, not for a reductive utilitarian calculus, but so as to be informed as to the diversity of opinions. Third, the phenomenon of diminishing marginal utility suggests that efforts should be directed to helping those who are worst off, rather than trying to make life better for the already well off, without, of course, presuming to automatically know what is best for others. Fourth, while we might disagree about the discount rate, the moral value of the future necessitates that we take downstream effects into account, rather than only focusing on immediate consequences. Sweeping attempts at regulation, such as GDPR, may have outsized effects here, as they will partially determine how we think about fairness going forward, and what it is legitimate to measure. Finally, because it is particularly difficult to predict consequences in the distant future, a high standard should be required for any policy that would place a definite burden on the present for a possible future gain.

\section{Randomization and Learning} \label{sec:algorithmic}

Before concluding, we will attempt to draw together a number of threads related to uncertainty, learning, and randomization. 
As described earlier, most philosophical presentations of consequentialism are highly abstract, without considering how one would practically determine what actions or rules are best. Given that statistics and machine learning arose specifically to deal with the problem of uncertainty, it is natural to ask whether there is any role for \emph{learning} in consequentialism.

Indeed, an entire subfield of machine learning exists precisely to deal with the problem of action selection in the face of uncertainty (so-called ``bandit'' problems, or reinforcement learning more broadly). As noted in the introduction, the reinforcement learning objective explicitly encodes the goal of maximizing some benefit over the long term. Algorithms designed to optimize this objective typically rely initially on random exploration to reduce uncertainty, thereby facilitating long-term ``exploitation'' of rewards. 

Not surprisingly, a number of papers have proposed using similar strategies as a way of achieving fair outcomes over the long-term.
For example, \citet{kroll.2017} suggest that adding randomness to hiring algorithms could help to debias them over time.
\citet{joseph.2016} consider the problem of learning a policy for making loans, and present an algorithm to do so without violating a particular notion of fairness.\footnote{In a companion paper, \citet{joseph.2016.rawls} proclaim their approach to be Rawlsian, but this seems to miss the key point of Rawls---namely, that we must account for inequalities due to circumstances (i.e., ``regardless of their initial place in the social system''; \citealp{rawls.1958}). Rather, the approach of \citet{joseph.2016} merely says we should learn to give loans to people who will best be able to pay them back.}  \citet{liu.2017} extend this work, again trying to satisfy fairness in the contextual bandit setting. Meanwhile, \citet{barabas.2018} suggest using randomization to facilitate causal inference about the ``social, structural, and psychological drivers'' of crime.

Randomization in decision making is a deep and important topic, and has been the focus of much past work in ethics \citep{lockwood.1983,freedman.1987,bird.2016,haushofer.2019}.
As noted above, it can be a source of fairness, if we take ``fair'' to mean that everyone deserves an equal chance. It may also be useful to prevent strategic manipulation of a system, and has a definite role in some parts of American law \citep{perry.2015,kroll.2017}.

Although temporal discounting in consequentialism is typically discussed in terms of present versus future value (e.g., helping people today versus investing in the future), a similar trade off applies to costly experimentation for the purpose of reducing future uncertainty. Indeed, this sort of approach has been widely adopted in industry in the form of A/B testing, as well as for adaptive trials in domains such as medicine \citep{lai.2015}. Moreover, there is clearly something appealing about the idea that it \emph{should} be morally incumbent upon people to improve their understanding of the world over time, not merely to act on their current understanding. However, randomization also raises a number of serious concerns.

First, as always, there is the problem of value, and the question of who gets to decide how to balance present costs against future benefits.
Second, there are good reasons to think that such an approach is unlikely to work in complex sociotechnical systems.
Although reinforcement learning has been extraordinarily successful in limited domains, such as game playing and online advertising, making reinforcement learning tractable generally requires assuming the existence of a stable environment, a limited space of actions, a clear reward signal, and a massive amount of training data. In most policy domains, we can expect to have none of these.  Third, there may be real costs associated with participation in such a process;  while a bank could conceivably choose to add randomness to a policy for granting loans (for the purpose of better learning who is likely to pay them back), 
giving loans to people who cannot afford them could have severe negative consequences for those individuals. 

There are clearly some domains where randomization is widely used, and seems well justified, especially from the perspective of consequentialism.
The best example of this is clinical trials in medicine, which are not only favored, but required. Medicine, however, is a special domain for several reasons: there is general agreement about ends (saving lives and reducing suffering), there is good reason to think that findings will generalize across people,
and there is a well-established framework for experimentation, with safeguards in place to protect the participants.

Where things get more complicated is using the same logic to establish the efficacy of social interventions, such as randomized trials in development economics.
Although controlled experiments do provide good evidence about whether an intervention was effective, it is less clear that the conclusions will generalize to different situations \citep{barrett.2010}.

Ultimately, while randomization can be an important tool in learning policies that promote long term benefits, especially in relatively static, generalizable domains, the limitations of both consequentialism and of statistical learning theory mean that we should be highly skeptical of any attempt to use it as the basis for creating policies or automated decision making systems to deal with complex social problems.

\section{Additional Related Work}

Beyond the criteria mentioned in section \S\ref{sec:fairness}, numerous other fairness metrics have been proposed, such as procedural fairness \citep{grgic.2016} and causal effects \citep{madras.2018,khademi.2019}.  Meanwhile, other papers have emphasized that simply satisfying a particular definition of fairness is no guarantee of the broader outcomes people care about, such as justice \citep{hu.2018}.
\citet{selbst.2019} discuss five common ``traps'' in thinking about sociotechnical systems, and \citet{friedler.2019} demonstrate how outcomes differs depending on preprocessing and the choice of fairness metric. 

Others have explored various types of consequences in particular settings, such as cost to the community in criminal justice \citep{corbett.2017}, runaway feedback loops in predictive policing \citep{ensign.2018}, disparities in the labour market \citep{hu.2018.www}, and the potential for strategic manipulation of policies \citep{hu.2019,milli.2019}.
\citet{liu.2018} demonstrate the importance of modeling the delayed impact of adopting various fairness metrics, even when focused narrowly on outcomes such as demographic parity.
In a discussion of racial bias in the criminal justice system, \citet{huq.2019} uses broadly consequentialist logic, arguing that the systems should be evaluated in terms of costs and benefits to minority groups.
For a discussion of Stoic philosophy in relation to AI, see \citet{murray.2017}.
%For a prescient early discussion of the connections between AI and utilitarianism, see \citet{good.1980}.
For surveys discussing the intersection of ethics and AI more broadly, see \citet{brundage.2014} and \citet{yu.2018}. For a book-length treatment of the subject, see \citet{wallach.2008}.

\section{Conclusions}

Consequentialism represents one of the most important pillars of ethical thinking in philosophy, including (but not limited to) utilitarianism. In brief, the central tenet of consequentialism is that actions should be evaluated in terms of the relative goodness of the expected outcomes, according to an impartial perspective on what is best. Despite  a number of serious problems that limit its practical application, including computational  problems involving value, uncertainty, and discounting, consequentialism still provides a useful basis for thinking about the limitations of other normative frameworks. 

Within the context of automated decision making, a consequentialist perspective underscores that merely satisfying a particular fairness metric is no guarantee of ethical conduct. Rather, consequentialism requires that we consider all possible options (including the possibility of not deploying an automated system), and weigh the likely consequences that will result, considered broadly, including possible implications for the long term future. Moreover, we must consider not only those who will be directly affected, but broader impacts on communities, and systemic effects of replacing many human decision makers with a single policy. While there are contexts in which it is reasonable, even required, to attempt to learn from the present for the benefit of the future, we should be skeptical of any randomization schemes which make unrealistic assumptions about the generalizability of what can be learned from social systems.

The political nature of valuation means we are unlikely to ever have agreement on what outcomes are best, and long term consequences will always remain to some extent unpredictable.
Nevertheless, through ongoing efforts to take into consideration a diverse set of perspectives on value, and systematic attempts to learn from our experiences, we can strive to move towards policies which are likely to lead to a better world, over both the short and long term future.

\section*{Acknowledgments}

The authors would like to thank Jared Moore, Emily Kalah Gade, Maarten Sap, Dan Hendrycks, Novi Quadrianto, and all reviewers for their thoughtful feedback and comments on this work.

\bibliography{refs_v4}

\begin{thebibliography}{77}
\providecommand{\natexlab}[1]{#1}
\providecommand{\url}[1]{\texttt{#1}}
\expandafter\ifx\csname urlstyle\endcsname\relax
  \providecommand{\doi}[1]{doi: #1}\else
  \providecommand{\doi}{doi: \begingroup \urlstyle{rm}\Url}\fi

\bibitem[Abel et~al.(2016)Abel, MacGlashan, and Littman]{abel.2016}
D.~Abel, J.~MacGlashan, and M.~L. Littman.
\newblock Reinforcement learning as a framework for ethical decision making.
\newblock In \emph{Proceedings of the Workshop on AI, Ethics, and Society at
  AAAI}, 2016.

\bibitem[Amodei et~al.(2016)Amodei, Olah, Steinhardt, Christiano, Schulman, and
  Man\'e]{amodei.2016}
D.~Amodei, C.~Olah, J.~Steinhardt, P.~Christiano, J.~Schulman, and D.~Man\'e.
\newblock Concrete problems in {AI} safety.
\newblock \emph{arXiv}, abs/1606.06565v2, 2016.

\bibitem[Anscombe(1958)]{anscombe.1958}
G.~E.~M. Anscombe.
\newblock Modern moral philosophy.
\newblock \emph{Philosophy}, 33\penalty0 (124), 1958.

\bibitem[Barabas et~al.(2018)Barabas, Virza, Dinakar, Ito, and
  Zittrain]{barabas.2018}
C.~Barabas, M.~Virza, K.~Dinakar, J.~Ito, and J.~Zittrain.
\newblock Interventions over predictions: Reframing the ethical debate for
  actuarial risk assessment.
\newblock In \emph{Proceedings of FAT*}, 2018.

\bibitem[Barocas and Selbst(2016)]{barocas.2016}
S.~Barocas and A.~D. Selbst.
\newblock Big data's disparate impact.
\newblock \emph{California Law Review}, 104, 2016.

\bibitem[Barrett and Carter(2010)]{barrett.2010}
C.~B. Barrett and M.~R. Carter.
\newblock The power and pitfalls of experiments in development economics: Some
  non-random reflections.
\newblock \emph{Applied Economic Perspectives and Policy}, 32\penalty0 (4),
  2010.

\bibitem[Benthall and Haynes(2019)]{benthall.2019}
S.~Benthall and B.~D. Haynes.
\newblock Racial categories in machine learning.
\newblock In \emph{Proceedings of FAT*}, 2019.

\bibitem[Bentham(1970 {[1781]})]{bentham}
J.~Bentham.
\newblock \emph{An Introduction to the Principles of Morals and Legislation}.
\newblock Oxford University Press, 1970 {[1781]}.

\bibitem[Bianco(2008)]{bianco.2008}
K.~M. Bianco.
\newblock \emph{The Subprime Lending Crisis: Causes and Effects of the Mortgage
  Meltdown}.
\newblock CCH, Wolters Kluwer Law \& Business, 2008.

\bibitem[Binmore(2009)]{binmore.2009}
K.~Binmore.
\newblock Interpersonal comparison of utility.
\newblock In D.~Ross and H.~Kincaid, editors, \emph{The Oxford Handbook of
  Philosophy of Economics}, chapter~20. Oxford University Press, 2009.

\bibitem[Binns(2018)]{binnis.2018}
R.~Binns.
\newblock Fairness in machine learning: Lessons from political philosophy.
\newblock In \emph{Proceedings of FAT*}, 2018.

\bibitem[Bird et~al.(2016)Bird, Barocas, Crawford, and Wallach]{bird.2016}
S.~Bird, S.~Barocas, K.~Crawford, and H.~Wallach.
\newblock Exploring or exploiting? {Social} and ethical implications of
  autonomous experimentation in {AI}.
\newblock In \emph{Proceedings of FAT/ML}, 2016.

\bibitem[Brundage(2014)]{brundage.2014}
M.~Brundage.
\newblock Limitations and risks of machine ethics.
\newblock \emph{Journal of Experimental \& Theoretical Artificial
  Intelligence}, 26, 2014.

\bibitem[Chouldechova(2017)]{chouldechova.2017}
A.~Chouldechova.
\newblock Fair prediction with disparate impact: a study of bias in recidivism
  prediction instruments.
\newblock \emph{arXiv}, abs/1703.00056, 2017.

\bibitem[Corbett-Davies et~al.(2017)Corbett-Davies, Pierson, Feller, Goel, and
  Huq]{corbett.2017}
S.~Corbett-Davies, E.~Pierson, A.~Feller, S.~Goel, and A.~Huq.
\newblock Algorithmic decision making and the cost of fairness.
\newblock In \emph{Proceedings of KDD}, 2017.

\bibitem[Cowen(2006)]{cowen.2006}
T.~Cowen.
\newblock The epistemic problem does not refute consequentialism.
\newblock \emph{Utilitas}, 18\penalty0 (4), 2006.

\bibitem[Cowen and Parfit(1992)]{cowen.1992}
T.~Cowen and D.~Parfit.
\newblock Against the social discount rate.
\newblock In P.~Laslett and J.~Fishkin, editors, \emph{Philosophy, Politics,
  and Society}. Yale University Press, 1992.

\bibitem[de~Lazari-Radek and Singer(2017)]{singer.2017}
K.~de~Lazari-Radek and P.~Singer.
\newblock \emph{Utilitarianism: A Very Short Introduction}.
\newblock Oxford University Press, 2017.

\bibitem[Driver(2005)]{driver.2005}
J.~Driver.
\newblock Consequentialism and feminist ethics.
\newblock \emph{Hypatia}, 20\penalty0 (4):\penalty0 183--199, 2005.

\bibitem[Driver(2012)]{driver.2012}
J.~Driver.
\newblock \emph{Consequentialism}.
\newblock Routledge, 2012.

\bibitem[Dwork et~al.(2012)Dwork, Hardt, Pitassi, Reingold, and
  Zemel]{dwork.2012}
C.~Dwork, M.~Hardt, T.~Pitassi, O.~Reingold, and R.~Zemel.
\newblock Fairness through awareness.
\newblock In \emph{Proceedings of ITCS}, 2012.

\bibitem[Ensign et~al.(2018)Ensign, Friedler, Neville, Scheidegger, and
  Venkatasubramanian]{ensign.2018}
D.~Ensign, S.~A. Friedler, S.~Neville, C.~Scheidegger, and
  S.~Venkatasubramanian.
\newblock Runaway feedback loops in predictive policing.
\newblock In \emph{Proceedings of FAT*}, 2018.

\bibitem[Foot(1967)]{foot.1967}
P.~Foot.
\newblock The problem of abortion and the doctrine of the double effect.
\newblock \emph{Oxford Review}, 5, 1967.

\bibitem[Freedman(1987)]{freedman.1987}
B.~Freedman.
\newblock Equipoise and the ethics of clinical research.
\newblock \emph{New England Journal of Medicine}, 317\penalty0 (3), 1987.

\bibitem[Friedler et~al.(2019)Friedler, Scheidegger, Venkatasubramanian,
  Choudhary, Hamilton, and Roth]{friedler.2019}
S.~A. Friedler, C.~Scheidegger, S.~Venkatasubramanian, S.~Choudhary, E.~P.
  Hamilton, and D.~Roth.
\newblock A comparative study of fairness-enhancing interventions in machine
  learning.
\newblock In \emph{Proceedings of FAT*}, 2019.

\bibitem[Friedman(1991)]{friedman.1991}
M.~Friedman.
\newblock The practice of partiality.
\newblock \emph{Ethics}, 101\penalty0 (4), 1991.

\bibitem[Green(2019)]{green.2019}
B.~Green.
\newblock ``{Good}'' isn't good enough.
\newblock In \emph{Proceedings of the AI for Social Good workshop at NeurIPS},
  2019.

\bibitem[Green and Hu(2018)]{green.2018}
B.~Green and L.~Hu.
\newblock The myth in the methodology: Towards a recontextualization of
  fairness in machine learning.
\newblock In \emph{Proceedings of the Debates workshop at ICML}, 2018.

\bibitem[Greene and Haidt(2002)]{greene.2002}
J.~Greene and J.~Haidt.
\newblock How (and where) does moral judgment work?
\newblock \emph{Trends in cognitive sciences}, 6\penalty0 (12), 2002.

\bibitem[Greene(2013)]{greene.2013}
J.~D. Greene.
\newblock \emph{Moral Tribes: Emotion, Reason, and the Gap between Us and
  Them}.
\newblock The Penguin Press, 2013.

\bibitem[{Grgi\'c-Hla\v ca} et~al.(2016){Grgi\'c-Hla\v ca}, Zafar, Gummadi, and
  Weller]{grgic.2016}
N.~{Grgi\'c-Hla\v ca}, M.~B. Zafar, K.~P. Gummadi, and A.~Weller.
\newblock The case for process fairness in learning: Feature selection for fair
  decision making.
\newblock In \emph{Proceedings of the Symposium on Machine Learning and the Law
  at NeurIPS}, 2016.

\bibitem[Hardt et~al.(2016)Hardt, Price, and Srebro]{hardt.2016}
M.~Hardt, E.~Price, and N.~Srebro.
\newblock Equality of opportunity in supervised learning.
\newblock In \emph{Proceedings of NeurIPS}, 2016.

\bibitem[Harsanyi(1977)]{harsanyi.1977}
J.~C. Harsanyi.
\newblock Rule utilitarianism and decision theory.
\newblock \emph{Erkenntnis}, 11:\penalty0 25--53, May 1977.

\bibitem[Haushofer et~al.(2019)Haushofer, Riis-Vestergaard, and
  Shapiro]{haushofer.2019}
J.~Haushofer, M.~I. Riis-Vestergaard, and J.~Shapiro.
\newblock Is there a social cost of randomization?
\newblock \emph{Social Choice and Welfare}, 52, 2019.

\bibitem[Hoffmann(2017)]{hoffmann.2017}
A.~L. Hoffmann.
\newblock Beyond distributions and primary goods: Assessing applications of
  {Rawls} in information science and technology literature since 1990.
\newblock \emph{Journal of the Association for Information Science and
  Technology}, 68\penalty0 (7), 2017.

\bibitem[Hooker(2002)]{hooker.2002}
B.~Hooker.
\newblock \emph{Ideal Code, Real World: A Rule-consequentialist Theory of
  Morality}.
\newblock Clarendon Press, 2002.

\bibitem[Hu and Chen(2018{\natexlab{a}})]{hu.2018}
L.~Hu and Y.~Chen.
\newblock Welfare and distributional impacts of fair classification.
\newblock In \emph{Proceedings of FAT/ML}, 2018{\natexlab{a}}.

\bibitem[Hu and Chen(2018{\natexlab{b}})]{hu.2018.www}
L.~Hu and Y.~Chen.
\newblock A short-term intervention for long-term fairness in the labor market.
\newblock In \emph{Proceedings of WWW}, 2018{\natexlab{b}}.

\bibitem[Hu et~al.(2019)Hu, Immorlica, and Vaughan]{hu.2019}
L.~Hu, N.~Immorlica, and J.~W. Vaughan.
\newblock The disparate effects of strategic manipulation.
\newblock In \emph{Proceedings of FAT*}, 2019.

\bibitem[Huq(2019)]{huq.2019}
A.~Z. Huq.
\newblock Racial equity in algorithmic criminal justice.
\newblock \emph{Duke Law Journal}, 68\penalty0 (6), 2019.

\bibitem[Joseph et~al.(2016{\natexlab{a}})Joseph, Kearns, Morgenstern, Neel,
  and Roth]{joseph.2016.rawls}
M.~Joseph, M.~Kearns, J.~H. Morgenstern, S.~Neel, and A.~Roth.
\newblock Rawlsian fairness for machine learning.
\newblock In \emph{Proceedings of FAT/ML}, 2016{\natexlab{a}}.

\bibitem[Joseph et~al.(2016{\natexlab{b}})Joseph, Kearns, Morgenstern, and
  Roth]{joseph.2016}
M.~Joseph, M.~Kearns, J.~H. Morgenstern, and A.~Roth.
\newblock Fairness in learning: Classic and contextual bandits.
\newblock In \emph{Proceedings of NeurIPS}, 2016{\natexlab{b}}.

\bibitem[Kagan(1991)]{kagan.1991}
S.~Kagan.
\newblock \emph{The Limits of Morality}.
\newblock Claredon Press, 1991.

\bibitem[Kagan(1998)]{kagan.1998}
S.~Kagan.
\newblock \emph{Normative Ethics}.
\newblock Westview Press, 1998.

\bibitem[Khademi et~al.(2019)Khademi, Lee, Foley, and Honavar]{khademi.2019}
A.~Khademi, S.~Lee, D.~Foley, and V.~Honavar.
\newblock Fairness in algorithmic decision making: An excursion through the
  lens of causality.
\newblock In \emph{Proceedings of WWW}, 2019.

\bibitem[Kittay(2009)]{kittay.2009}
E.~F. Kittay.
\newblock The ethics of philosophizing: Ideal theory and the exclusion of
  people with severe cognitive disabilities.
\newblock In L.~Tessman, editor, \emph{Feminist Ethics and Social and Political
  Philosophy: Theorizing the Non-Ideal}, chapter~8, pages 121--146. Springer,
  2009.

\bibitem[Kleinberg et~al.(2017)Kleinberg, Mullainathan, and
  Raghavan]{kleinberg.2017}
J.~M. Kleinberg, S.~Mullainathan, and M.~Raghavan.
\newblock Inherent trade-offs in the fair determination of risk scores.
\newblock In \emph{Proceedings of ITCS}, 2017.

\bibitem[Kleinberg et~al.(2018)Kleinberg, Ludwig, Mullainathan, and
  Rambachan]{kleinberg.2018}
J.~M. Kleinberg, J.~Ludwig, S.~Mullainathan, and A.~Rambachan.
\newblock Algorithmic fairness.
\newblock In \emph{AEA Papers and Proceedings}, 2018.

\bibitem[Kleindessner et~al.(2019)Kleindessner, Samadi, Awasthi, and
  Morgenstern]{kleindessner.2019}
M.~Kleindessner, S.~Samadi, P.~Awasthi, and J.~Morgenstern.
\newblock Guarantees for spectral clustering with fairness constraints.
\newblock In \emph{Proceedings of ICML}, 2019.

\bibitem[Kroll et~al.(2017)Kroll, Barocas, Felten, Reidenberg, Robinson, and
  Yu]{kroll.2017}
J.~A. Kroll, S.~Barocas, E.~W. Felten, J.~R. Reidenberg, D.~G. Robinson, and
  H.~Yu.
\newblock Accountable algorithms.
\newblock \emph{University of Pennsylvania Law Review}, 165\penalty0 (3), 2017.

\bibitem[Lai et~al.(2015)Lai, Lavori, and Tsang]{lai.2015}
T.~L. Lai, P.~W. Lavori, and K.~W. Tsang.
\newblock Adaptive design of confirmatory trials: Advances and challenges.
\newblock \emph{Contemporary clinical trials}, 45:\penalty0 93--102, 2015.

\bibitem[Liu et~al.(2018)Liu, Dean, Rolf, Simchowitz, and Hardt]{liu.2018}
L.~T. Liu, S.~Dean, E.~Rolf, M.~Simchowitz, and M.~Hardt.
\newblock Delayed impact of fair machine learning.
\newblock In \emph{Proceedings of ICML}, 2018.

\bibitem[Liu et~al.(2017)Liu, Radanovic, Dimitrakakis, Mandal, and
  Parkes]{liu.2017}
Y.~Liu, G.~Radanovic, C.~Dimitrakakis, D.~Mandal, and D.~C. Parkes.
\newblock Calibrated fairness in bandits.
\newblock In \emph{Proceedings of FAT/ML}, 2017.

\bibitem[Lockwood and Anscombe(1983)]{lockwood.1983}
M.~Lockwood and G.~E.~M. Anscombe.
\newblock Sins of omission? {The} non-treatment of controls in clinical trials.
\newblock \emph{Aristotelian Society Supplementary Volume}, 57\penalty0 (1),
  1983.

\bibitem[Madras et~al.(2018)Madras, Pitassi, and Zemel]{madras.2018}
D.~Madras, T.~Pitassi, and R.~Zemel.
\newblock Predict responsibly: Improving fairness and accuracy by learning to
  defer.
\newblock In \emph{Proceedings of NeurIPS}, 2018.

\bibitem[Mill(1979 {[1863]})]{mill}
J.~S. Mill.
\newblock \emph{Utilitarianism}.
\newblock Hackett Publishing Company, Inc., 1979 {[1863]}.

\bibitem[Milli et~al.(2019)Milli, Miller, Dragan, and Hardt]{milli.2019}
S.~Milli, J.~Miller, A.~D. Dragan, and M.~Hardt.
\newblock The social cost of strategic classification.
\newblock In \emph{Proceedings of FAT*}, 2019.

\bibitem[Mills(1987)]{mills.1997}
C.~W. Mills.
\newblock \emph{The Racial Contract}.
\newblock Cornell University Press, 1987.

\bibitem[Murray(2017)]{murray.2017}
G.~Murray.
\newblock Stoic ethics for artificial agents.
\newblock \emph{arXiv}, abs/1701.02388, 2017.

\bibitem[Parfit(1984)]{parfit}
D.~Parfit.
\newblock \emph{Reasons and Persons}.
\newblock Oxford University Press, 1984.

\bibitem[Perry and Zarsky(2015)]{perry.2015}
R.~Perry and T.~Zarsky.
\newblock ``{M}ay the odds be ever in your favour'': Lotteries in law.
\newblock \emph{Alabama Law Review}, 66\penalty0 (5):\penalty0 1035--1098,
  2015.

\bibitem[Portmore(2011)]{portmore.2011}
D.~W. Portmore.
\newblock \emph{Commonsense Consequentialism}.
\newblock Oxford University Press, 2011.

\bibitem[Railton(1984)]{railton}
P.~Railton.
\newblock Alientation, consequentailism, and the demands of moraltiy.
\newblock \emph{Philosophy and Public Affairs}, 13, 1984.

\bibitem[Rawls(1958)]{rawls.1958}
J.~Rawls.
\newblock Justice as fairness.
\newblock \emph{Philosophical Review}, LXVII, 1958.

\bibitem[Rawls(1971)]{rawls.1971}
J.~Rawls.
\newblock \emph{A Theory of Justice}.
\newblock The Belknap Press of Harvard University, 1971.

\bibitem[Ruggieri et~al.(2010)Ruggieri, Pedreschi, and Turini]{ruggieri.2010}
S.~Ruggieri, D.~Pedreschi, and F.~Turini.
\newblock Data mining for discrimination discovery.
\newblock \emph{ACM Trans. Knowl. Discov. Data}, 4\penalty0 (2), 2010.

\bibitem[Scanlon(1998)]{scanlon.1998}
T.~M. Scanlon.
\newblock \emph{What We Owe to Each Other}.
\newblock Harvard University Press, 1998.

\bibitem[Scheffler(1994)]{scheffler}
S.~Scheffler.
\newblock \emph{The Rejection of Consequentialism: A Philosophical
  Investigation of the Considerations Underlying Rival Moral Conceptions}.
\newblock Oxford University Press, 1994.

\bibitem[Schultz and Varouxakis(2005)]{schultz.2005}
B.~Schultz and G.~Varouxakis, editors.
\newblock \emph{Utilitarianism and Empire}.
\newblock Lexington Books, 2005.

\bibitem[Selbst et~al.(2019)Selbst, boyd d., Friedler, Venkatasubramanian, and
  Vertesi]{selbst.2019}
A.~D. Selbst, boyd d., S.~A. Friedler, S.~Venkatasubramanian, and J.~Vertesi.
\newblock Fairness and abstraction in sociotechnical systems.
\newblock In \emph{Proceedings of FAT*}, 2019.

\bibitem[Sidgwick(1967)]{sidgwick}
H.~Sidgwick.
\newblock \emph{The Method of Ethics}.
\newblock Macmillon, 1967.

\bibitem[Singer(1972)]{singer.1972}
P.~Singer.
\newblock Famine, affluence, and morality.
\newblock \emph{Philosophy and Public Affairs}, 1\penalty0 (1):\penalty0
  229--243, 1972.

\bibitem[Singer(1993)]{singer.1993}
P.~Singer.
\newblock \emph{Practical Ethics}.
\newblock Cambridge University Press, 1993.

\bibitem[Smart and Williams(1973)]{smart.1973}
J.~J.~C. Smart and B.~Williams.
\newblock \emph{Utilitarianism: For \& Against}.
\newblock Cambridge University Press, 1973.

\bibitem[Torresen(2018)]{torresen.2018}
J.~Torresen.
\newblock A review of future and ethical perspectives of robotics and {AI}.
\newblock \emph{Frontiers in Robotics and AI}, 4, 2018.

\bibitem[Wallach and Allen(2008)]{wallach.2008}
W.~Wallach and C.~Allen.
\newblock \emph{Moral Machines: {Teaching} Robots Right from Wrong}.
\newblock Oxford University Press, Inc., 2008.

\bibitem[Yu et~al.(2018)Yu, Shen, Miao, Leung, Lesser, and Yang]{yu.2018}
H.~Yu, Z.~Shen, C.~Miao, C.~Leung, V.~R. Lesser, and Q.~Yang.
\newblock Building ethics into artificial intelligence.
\newblock In \emph{Proceedings of IJCAI}, 2018.

\end{thebibliography}

\end{document}